\documentclass{article}

\usepackage{microtype}
\usepackage{graphicx}
\usepackage{subfigure}
\usepackage{booktabs} 
\usepackage{tikz}
\usepackage{tabularx}
\usepackage{multirow}
\usepackage{amsmath}
\usepackage{amssymb}
\usepackage{dsfont}
\usepackage{bm}

\usepackage{hyperref}

\usepackage{inconsolata}
\usepackage[noabbrev,capitalize]{cleveref}

\usepackage[accepted]{icml2020}

\newcommand{\MYhref}[3][blue]{\href{#2}{\color{#1}{#3}}}%
\icmltitlerunning{Regional Image Perturbation Reduces $L_p$ Norms of Adversarial Examples While Maintaining Model-to-model Transferability}

\begin{document}

\twocolumn[

\icmltitle{Regional Image Perturbation Reduces $L_p$ Norms of Adversarial Examples While Maintaining Model-to-model Transferability}




\icmlsetsymbol{equal}{*}

\begin{icmlauthorlist}
\icmlauthor{Utku Ozbulak}{equal,a,f}
\icmlauthor{Jonathan Peck}{equal,c,d}
\icmlauthor{Wesley De Neve}{a,f}
\icmlauthor{Bart Goossens}{e}\\
\icmlauthor{Yvan Saeys}{c,d}
\icmlauthor{Arnout Van Messem}{c,f}
\end{icmlauthorlist}

\icmlaffiliation{a}{Department of Electronics and Information Systems, Ghent University, Ghent, Belgium}
\icmlaffiliation{c}{Department of Applied Mathematics, Computer Science and Statistics, Ghent University, Ghent, Belgium}
\icmlaffiliation{d}{Data Mining and Modeling for Biomedicine, VIB Inflammation Research Center, Ghent, Belgium}
\icmlaffiliation{e}{Department of Telecommunications and Information Processing, Ghent University - imec, Ghent, Belgium}
\icmlaffiliation{f}{Center for Biotech Data Science, Ghent University Global Campus, Incheon, Republic of Korea}

\icmlcorrespondingauthor{Utku Ozbulak}{utku.ozbulak@ugent.be}

\icmlkeywords{Regional image perturbation, Adversarial Examples, Adversarial Attacks, Robustness}

\vskip 0.3in
]



\printAffiliationsAndNotice{\icmlEqualContribution}

\begin{abstract}
Regional adversarial attacks often rely on complicated methods for generating adversarial perturbations, making it hard to compare their efficacy against well-known attacks. In this study, we show that effective regional perturbations can be generated without resorting to complex methods. We develop a very simple regional adversarial perturbation attack method using cross-entropy sign, one of the most commonly used losses in adversarial machine learning. Our experiments on ImageNet with multiple models reveal that, on average, $76\%$ of the generated adversarial examples maintain model-to-model transferability when the perturbation is applied to local image regions. Depending on the selected region, these localized adversarial examples require significantly less $L_p$ norm distortion (for $p \in \{0, 2, \infty\}$) compared to their non-local counterparts. These localized attacks therefore have the potential to undermine defenses that claim robustness under the aforementioned norms. 
\vspace{-1em}
\end{abstract}

\section{Introduction}
\label{Introduction}

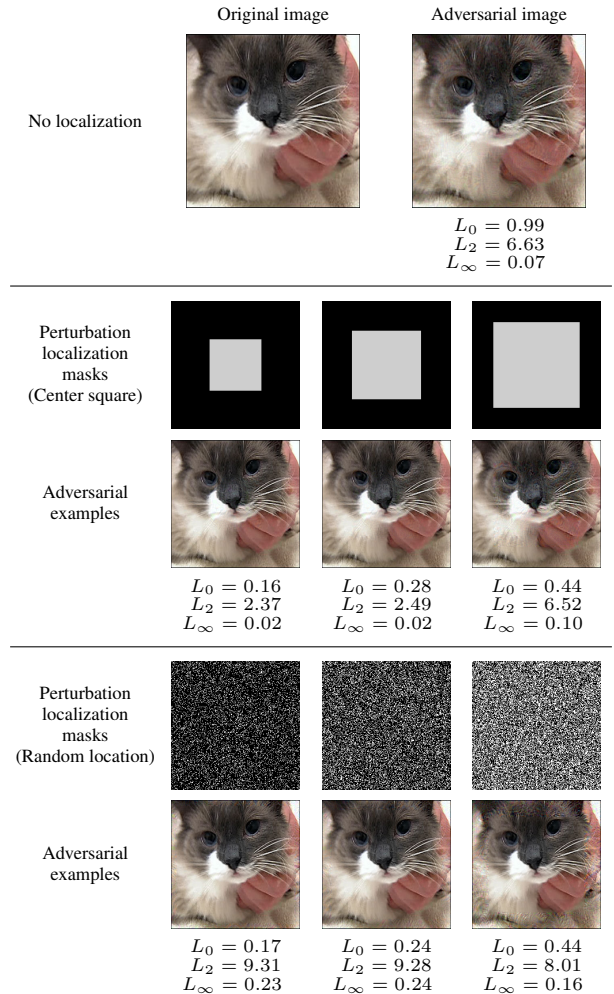
\begin{figure}[h!]
\centering
\begin{tikzpicture}
\scriptsize  
\centering
\def\sety1{0}
\node[align=center] at (0, \sety1) {No localization};
\node[inner sep=0pt] (russell) at (2.5, \sety1)
    {\includegraphics[width=2.3cm]{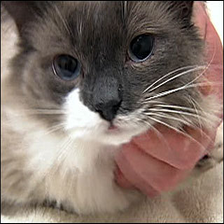}};
\node[align=center] at (2.5, \sety1 + 1.4) {Original image};
\node[inner sep=0pt] (whitehead) at (5.5, \sety1)
    {\includegraphics[width=2.3cm]{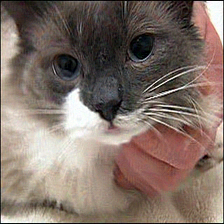}};
\node[align=center] at (5.5, \sety1 + 1.4) {Adversarial image};
\node[align=center] at (5.5, \sety1 - 1.4) {$L_0=0.99$};
\node[align=center] at (5.5, \sety1 - 1.65) {$L_2=6.63$};
\node[align=center] at (5.5-0.05, \sety1 - 1.9) {$L_{\infty}=0.07$};

\draw (-1, \sety1 -2.2) -- (7, \sety1 -2.2);

\def\sety1{-3.25}
\node[align=center] at (0, \sety1) {Perturbation\\localization\\masks\\(Center square)};
\node[inner sep=0pt] (russell) at (2, \sety1)
    {\includegraphics[width=1.7cm]{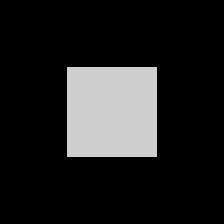}};
\node[inner sep=0pt] (whitehead) at (4, \sety1)
    {\includegraphics[width=1.7cm]{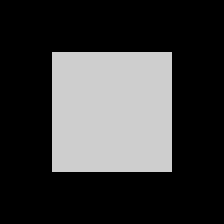}};
\node[inner sep=0pt] (whitehead) at (6, \sety1)
    {\includegraphics[width=1.7cm]{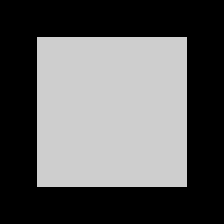}};

\def\sety1{-5.1}
\node[align=center] at (0, \sety1) {Adversarial\\examples};
\node[inner sep=0pt] (russell) at (2, \sety1)
    {\includegraphics[width=1.7cm]{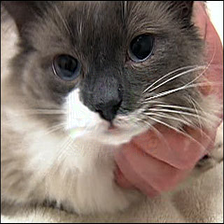}};
\node[align=center] at (2, \sety1-1.1) {$L_0=0.16$};
\node[align=center] at (2, \sety1-1.35) {$L_2=2.37$};
\node[align=center] at (2-0.05, \sety1-1.6) {$L_{\infty}=0.02$};
\node[inner sep=0pt] (whitehead) at (4, \sety1)
    {\includegraphics[width=1.7cm]{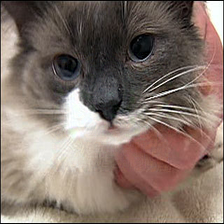}};
\node[align=center] at (4, \sety1-1.1) {$L_0=0.28$};
\node[align=center] at (4, \sety1-1.35) {$L_2=2.49$};
\node[align=center] at (4-0.05, \sety1-1.6) {$L_{\infty}=0.02$};
\node[inner sep=0pt] (whitehead) at (6, \sety1)
    {\includegraphics[width=1.7cm]{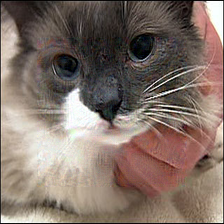}};
\node[align=center] at (6, \sety1-1.1) {$L_0=0.44$};
\node[align=center] at (6, \sety1-1.35) {$L_2=6.52$};
\node[align=center] at (6-0.05, \sety1-1.6) {$L_{\infty}=0.10$};

\draw (-1, \sety1 -1.9) -- (7, \sety1 -1.9);

\def\sety1{-8.05}
\node[align=center] at (0, \sety1) {Perturbation\\localization\\masks\\(Random location)};
\node[inner sep=0pt] (russell) at (2, \sety1)
    {\includegraphics[width=1.7cm]{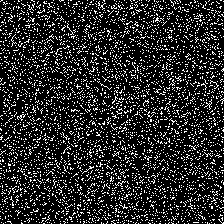}};
\node[inner sep=0pt] (whitehead) at (4, \sety1)
    {\includegraphics[width=1.7cm]{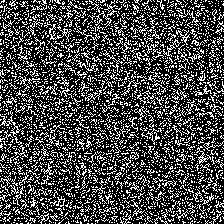}};
\node[inner sep=0pt] (whitehead) at (6, \sety1)
    {\includegraphics[width=1.7cm]{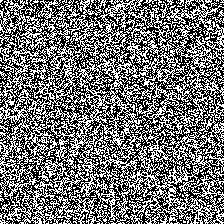}};

\def\sety1{-9.9}
\node[align=center] at (0, \sety1) {Adversarial\\examples};
\node[inner sep=0pt] (russell) at (2, \sety1)
    {\includegraphics[width=1.7cm]{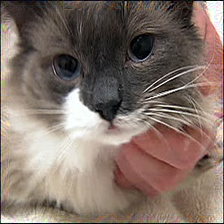}};
\node[align=center] at (2, \sety1-1.1) {$L_0=0.17$};
\node[align=center] at (2, \sety1-1.35) {$L_2=9.31$};
\node[align=center] at (2-0.05, \sety1-1.6) {$L_{\infty}=0.23$};
\node[inner sep=0pt] (whitehead) at (4, \sety1)
    {\includegraphics[width=1.7cm]{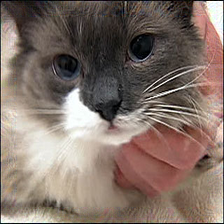}};
\node[align=center] at (4, \sety1-1.1) {$L_0=0.24$};
\node[align=center] at (4, \sety1-1.35) {$L_2=9.28$};
\node[align=center] at (4-0.05, \sety1-1.6) {$L_{\infty}=0.24$};
\node[inner sep=0pt] (whitehead) at (6, \sety1)
    {\includegraphics[width=1.7cm]{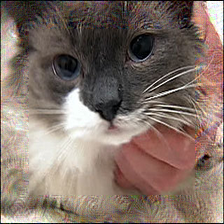}};
\node[align=center] at (6, \sety1-1.1) {$L_0=0.44$};
\node[align=center] at (6, \sety1-1.35) {$L_2=8.01$};
\node[align=center] at (6-0.05, \sety1-1.6) {$L_{\infty}=0.16$};
\end{tikzpicture}\vspace{-1.2em}
\caption{(Top) An input image and its adversarial counterpart created with IFGS. (Center and bottom) Perturbation localization grids illustrated with black-gray images and adversarial examples generated by IFGS when the perturbation is only applied to the grey areas in the localization grids. $L_p$ norms of the perturbation are provided under each image. All of the adversarial examples were generated using AlexNet and successfully transfer to ResNet-50.}
\label{fig:adv_sq_fr}
\vspace{-1em}
\end{figure}

Recent advancements in the field of machine learning (ML) --- more specifically, in deep learning (DL) --- have substantially increased the adoption rate of automated systems in everyday life~\cite{Alexnet,resnet,resnext}. However, since their inception, these systems have been criticized for their lack of \textit{interpretability}: it is often difficult or impossible to know precisely \textit{why} an ML model produces a specific response for a given input, yet such information is highly relevant in many settings~\cite{interpretation_fragile,saliency_unreliability}. One manifestation of this shortcoming of current ML theory to understand DL models is the phenomenon of \textit{adversarial examples}~\cite{LBFGS}, which has recently received much attention in the research community. Adversarial examples are data points specifically crafted by an adversary in order to force machine learning models into making mistakes. Often, these artificial examples are visually indistinguishable from natural data points, making it almost impossible for humans to detect them and calling into question the generalization ability of deep neural networks (DNN)~\cite{schmidt2018adversarially,ilyas2019adversarial}.


Formally, adversarial examples are usually defined as follows~\cite{LBFGS,PGD_attack}. Given an ML model $f$ and an input $\bm{X}$, an adversarial example $\tilde{\bm{X}}$ satisfies (1) $\|\bm{X} - \tilde{\bm{X}}\|_p \leq \varepsilon$ for some chosen $L_p$ norm and perturbation budget $\varepsilon > 0$ and (2) $f(\bm{X}) \neq f(\tilde{\bm{X}})$. In other words, the perturbed input $\tilde{\bm{X}}$ must be ``close'' to the original input $\bm{X}$ as measured by an $L_p$ norm and the classifier $f$ must output different labels for $\bm{X}$ and $\tilde{\bm{X}}$. However, for sufficiently small values of $\varepsilon$, the two inputs are indistinguishable and should belong to the same class. Hence, the existence of adversarial examples for very small perturbation budgets indicates a failure of DL models to accurately capture the data manifold. Interestingly, depending on the attack used, adversarial examples can be highly \textit{transferable}: an adversarial sample $\tilde{\bm{X}}$ that fools a certain classifier $f$ can also fool completely different classifiers trained for the same task~\cite{DBLP:journals/corr/PapernotMG16,cheng2019improvinge_black_black_box}. This so-called \textit{transferability}, i.e., the degree to which an adversarial sample can fool other models, is a popular metric for assessing the effectiveness of a particular attack.


Through the generations of research in computer vision it was established that certain regions of images are more important for the identification of an object of interest than others~\citep{moravev_image_parts_important,schmid1997local_image_parts_important,lowe_SIFT,sun2014deep_face_parts_important,guided_backprop,grad_cam}. As such, research on localized adversarial attacks also shows that adversarial perturbation applied to these \textit{important} regions may change the prediction faster and with less $L_p$ perturbation than attacks that apply the perturbation to the entire image~\cite{one_pixel_attack,LAVAN,xu2018structured_new_local_adv_attack,zajac2019adversarial_frame}. However, analyses to prevent adversarial examples often do not evaluate robustness against such regional attacks. Adversarial defenses are often studied exclusively against well-understood attacks such as FGS~\cite{Goodfellow-expharnessing}, JSMA~\cite{JSMA}, IFGS~\cite{IFGS}, Carlini \& Wagner's Attack~\cite{CW_Attack}, PGD~\cite{PGD_attack}, and BPDA~\cite{athalye2018_obfuscated}, where these attacks apply their perturbations to the entire image based on the magnitude of the loss gradient for each pixel and according to the $L_p$ norm constraints they set. We believe this lack of evaluation against regional attacks is because (1) regional attacks are often studied in permissive white-box settings which do not represent real-world scenarios and (2) the proposed attacks usually come with a completely new and complicated way of generating adversarial examples, thus making it not straightforward to apply these attacks to different datasets, especially not locally, as opposed to well-understood attacks.


In this work, we show that so-called ``global'' adversarial attacks can be easily modified to become localized attacks. As such, different from previous research efforts on localized perturbation, our study does not propose a novel attack. Instead, we introduce a general method for localizing the perturbations generated by existing non-localized attacks. We achieve this by multiplying the original perturbations by a simple binary mask (as shown in \cref{fig:adv_sq_fr}), restricting the perturbation to specific image regions. We analyze both the transferability and the $L_p$ norm properties of the crafted adversarial examples, finding that the localized examples are about as effective as the examples generated by the original attacks (in terms of transferability), and with the localized versions often requiring significantly less $L_p$ distortion. The implementation of the proposed method is publicly available.\footnote{\texttt{ \MYhref[magenta]{https://github.com/utkuozbulak/regional-adversarial-perturbation}{github.com/utkuozbulak/\\regional-adversarial-perturbation}}}

The finding that we can significantly reduce the required $L_p$ distortion while maintaining similar levels of effectiveness potentially undermines many existing defenses --- certified or not --- since these usually guarantee robustness against specific $L_p$ perturbation budgets~\cite{wong2017provable,croce2018provable,andriushchenko2019provably,ghiasi2020breaking_certified_defenses}. Reducing the required distortion attacks below such thresholds could render these defenses ineffective.



\section{Framework}
\label{Framework}

\textbf{Data} Adversarial examples are mainly studied on MNIST~\cite{lecun1998gradient}, CIFAR~\cite{CIFAR}, or ImageNet~\cite{ILSVRC15:rus}. However, as the field of adversarial machine learning evolved, due to shortcomings in terms of color channels and image sizes, the MNIST and CIFAR datasets are not deemed to be suitable for studies that represent real-world scenarios where adversarial examples pose a threat~\cite{DBLP:journalsCarliniW17}. Following this observation, we use images taken from the test set of the ImageNet dataset in order to generate adversarial examples.

\textbf{Models} Although convolutional architectures were already used in the work of~\citet{lecun1998gradient}, it was the success of AlexNet in 2012 that popularized DNN architectures~\cite{Alexnet}. Recent research in the field of adversarial robustness also revealed AlexNet to be one of the more robust architectures~\cite{su2018robustness_18_imagenet_models_evaluation}. Following the success of AlexNet, VGG~\cite{VGG} architectures were proposed with smaller convolutional kernel sizes. Thanks to their simple architecture, VGG architectures are still popular today in many computer vision approaches. In order to overcome problems with vanishing gradients in deep architectures,~\citet{resnet} proposed ResNet architectures, introducing the usage of residual layers. These residual architectures were later expanded upon and are currently some of the most frequently used architectures for solving a variety of problems in the field of deep learning~\cite{resnext,resnext_video}. Given the history of the aforementioned architectures in the field of adversarial machine learning, as well as in other deep learning areas, we opted for the use of AlexNet, VGG-16, and ResNet-50 in our experiments.

\section{Experimental Setup}
\label{Experimental Setup}
\textbf{Generating adversarial examples} \citet{DBLP:journalsCarliniW17} demonstrated the fragility of adversarial examples generated by single-step attacks and argued that iterative attacks should be used for evaluating novel defenses. Iterative attacks calculate and add perturbation to the input iteratively according to the rule
\begin{align}
    \bm{X}_{n+1} = \bm{X}_{n} + \bm{P}_{n} \,,
\end{align}
where $\bm{X}_{n}$ and $\bm{P}_{n}$ represent the input and the perturbation generated at the $n$th iteration, respectively. In this study we generate the perturbation as follows:
\begin{align}
&\bm{P}_{n} = \alpha \, \text{sign} \big( \nabla_x J(g(\theta, \bm{X}_{n})_c) \big) \,,
\end{align}
where $\nabla_x J(g(\theta, \bm{X}_{n})_c)$ represents the gradient with respect to $\bm{X}$ obtained with the cross-entropy loss ($J$) when targeting the class $c$. We use $\alpha=0.004$ as the perturbation multiplier which approximately corresponds to changing the pixel values of images by $1/255$ at each iteration and perform this attack for $250$ iterations. Typically, adversarial attacks such as FGS, IFGS, and PGD enforce a constraint on the magnitude $\|\bm{X} - \tilde{\bm{X}}\|_p$ of the perturbation. However, in order to make a valid comparison between adversarial examples in terms of $L_0$, $L_{2}$, and $L_{\infty}$ norms, we only enforce a discretization constraint, thus ensuring that the produced adversarial examples can be represented as valid images (i.e., the pixel values of $\tilde{\bm{X}}$ lie within the range $[0,1]$, as can be expected from regular images).

\textbf{Localizing adversarial perturbation} In a previous research effort, we successfully used the Hadamard product to select target pixels for generating adversarial examples in the context of semantic segmentation~\cite{ozbulak2019impact}. In order to localize the perturbation to selected regions, we employ a similar approach in this research effort, making use of
\begin{align}
    \bm{X}_{n+1} = \bm{X}_{n} + \bm{P}_{n} \odot \bm{L} \,,
\end{align}
where $\bm{L}$ is a \textit{localization mask}, i.e., a binary tensor of the same shape as the input. In this tensor, regions where the perturbation needs to be applied are set to $1$ while the remainder is set to $0$.

\begin{figure*}
\setlength\unitlength{1cm}
\centering
\begin{minipage}{.3\textwidth}
\centering
\begin{picture}(4,4)
\multiput(0.1,0.1)(0,1){4}{\line(1,0){3}}
\multiput(0.1,0.1)(1,0){4}{\line(0,1){3}}
\begin{scriptsize}
\put(0.3,0.55){$52\%$}
\put(1.3,0.55){$59\%$}
\put(2.25,0.55){$100\%$}

\put(0.3,1.55){$60\%$}
\put(1.25,1.55){$100\%$}
\put(2.3,1.55){$56\%$}

\put(0.25,2.55){$100\%$}
\put(1.3,2.55){$73\%$}
\put(2.3,2.55){$66\%$}

\put(-0.3,2.3){\rotatebox{90}{AlexNet}}
\put(-0.3,1.25){\rotatebox{90}{VGG-16}}
\put(-0.3,0.05){\rotatebox{90}{ResNet-50}}

\put(0.12,3.3){AlexNet}
\put(1.12,3.3){VGG-16}
\put(2.12,3.3){ResNet-50}
\end{scriptsize}
\put (-0.75,0.75){\rotatebox{90}{Source Model}}
\put(0.75,3.75){Target Model}
\put(-0.5,-0.5){(a) $17\%$ of pixels selected}
\end{picture}
\end{minipage}
\begin{minipage}{.3\textwidth}
\centering
\begin{picture}(4,4)
\multiput(0.1,0.1)(0,1){4}{\line(1,0){3}}
\multiput(0.1,0.1)(1,0){4}{\line(0,1){3}}
\begin{scriptsize}
\put(0.3,0.55){$67\%$}
\put(1.3,0.55){$65\%$}
\put(2.25,0.55){$100\%$}

\put(0.3,1.55){$70\%$}
\put(1.25,1.55){$100\%$}
\put(2.3,1.55){$63\%$}

\put(0.25,2.55){$100\%$}
\put(1.3,2.55){$76\%$}
\put(2.3,2.55){$67\%$}

\put(-0.3,2.3){\rotatebox{90}{AlexNet}}
\put(-0.3,1.25){\rotatebox{90}{VGG-16}}
\put(-0.3,0.05){\rotatebox{90}{ResNet-50}}

\put(0.12,3.3){AlexNet}
\put(1.12,3.3){VGG-16}
\put(2.12,3.3){ResNet-50}
\end{scriptsize}
\put (-0.75,0.75){\rotatebox{90}{Source Model}}
\put(0.75,3.75){Target Model}
\put(-0.5,-0.5){(b) $28\%$ of pixels selected}
\end{picture}
\end{minipage}
\begin{minipage}{.3\textwidth}
\centering
\begin{picture}(4,4)
\multiput(0.1,0.1)(0,1){4}{\line(1,0){3}}
\multiput(0.1,0.1)(1,0){4}{\line(0,1){3}}
\begin{scriptsize}
\put(0.3,0.55){$78\%$}
\put(1.3,0.55){$77\%$}
\put(2.25,0.55){$100\%$}

\put(0.3,1.55){$83\%$}
\put(1.25,1.55){$100\%$}
\put(2.3,1.55){$75\%$}

\put(0.25,2.55){$100\%$}
\put(1.3,2.55){$78\%$}
\put(2.3,2.55){$75\%$}

\put(-0.3,2.3){\rotatebox{90}{AlexNet}}
\put(-0.3,1.25){\rotatebox{90}{VGG-16}}
\put(-0.3,0.05){\rotatebox{90}{ResNet-50}}

\put(0.12,3.3){AlexNet}
\put(1.12,3.3){VGG-16}
\put(2.12,3.3){ResNet-50}
\end{scriptsize}
\put (-0.75,0.75){\rotatebox{90}{Source Model}}
\put(0.75,3.75){Target Model}
\put(-0.5,-0.5){(c) $45\%$ of pixels selected}
\end{picture}
\end{minipage}
\vspace{1em}
\caption{Percentage of adversarial examples with localized perturbation that transfer from source model (generated from) to target model (tested against) when 17\%, 28\%, and 45\% of pixels are selected, respectively (combining all 3 localization approaches).}
\label{fig:localized_transferability}
\vspace{-1em}
\end{figure*}   

\begin{table*}[t]
\centering
\caption{Mean (standard deviation) $L_{\{2, \infty\}}$ distances calculated between genuine images and their adversarial counterparts, with the adversarial counterparts created by localization of perturbation (see the first column). Adversarial examples are created from the source models listed in the first row and transfer to the target models listed in the second row.}
\tiny
\begin{tabular}{llcc|cc|cc|cc|cc|cc}
	\cmidrule[1pt]{1-14}
	 \multirow{3}{*}{\rotatebox[origin=c]{90}{Localization}} & Source: & \multicolumn{4}{c}{AlexNet} & \multicolumn{4}{c}{VGG-16} & \multicolumn{4}{c}{ResNet-50}\\ 
    \cmidrule[0.25pt]{4-5} \cmidrule[0.25pt]{8-9} \cmidrule[0.25pt]{12-13} 
    ~ & Target: & \multicolumn{2}{c}{VGG-16} & \multicolumn{2}{c}{ResNet-50} & \multicolumn{2}{c}{AlexNet} & \multicolumn{2}{c}{ResNet-50} & \multicolumn{2}{c}{AlexNet} & \multicolumn{2}{c}{VGG-16} \\ 
    \cmidrule[0.25pt]{3-14}
    ~ & Norm: & $L_2$ & $L_{\infty}$ & $L_2$ & $L_{\infty}$ & $L_2$ & $L_{\infty}$ & $L_2$ & $L_{\infty}$ & $L_2$ & $L_{\infty}$ &  $L_2$ & $L_{\infty}$ \\ 
        \cmidrule[0.5pt]{1-14} 
        \multicolumn{2}{c}{\multirow{2}{*}{No Localization}} & $7.35$ 	& $0.07$	& $6.39$ 	& $0.05$	& $6.91$ 	& $0.07$ 	& $3.62$ 	& $0.02$ 	& $6.79$ 	& $0.07$	& $3.76$ 	& $0.02$ \\
        ~	& ~	&  $(5.37)$ 	& $(0.08)$ 	& $(4.50)$ 	& $(0.06)$ &  $(4.17)$ 	& $(0.05)$ 	& $(3.16)$ 	& $(0.04)$  	& $(4.31)$ 	& $(0.06)$ 	& $(2.20)$ 	& $(0.02)$\\
        \cmidrule[0.25pt]{1-14}
        \multirow{6}{*}{\rotatebox[origin=c]{90}{Center}} & \multirow{2}{*}{$90$px}	& $6.55$ 	& $0.15$ 	& $5.33$ 	& $0.11$ 	& $4.01$ 	& $0.10$	& $3.41$ 	& $0.09$ 	& $3.54$ 	& $0.09$ 	& $2.79$ 	& $0.06$ \\
        ~	& ~	& $(4.36)$ 	& $(0.14)$ 	& $(3.52)$ 	& $(0.10)$ 	& $(2.64)$ 	& $(0.09)$ 	& $(2.77)$ 	& $(0.10)$ 	& $(2.54)$ 	& $(0.09)$ 	& $(2.23)$ 	& $(0.07)$\\
        ~ & \multirow{2}{*}{$120$px}	& $6.47$ 	& $0.11$ 	& $6.30$ 	& $0.10$ 	& $5.01$ 	& $0.10$ 	& $3.68$ 	& $0.06$ 	& $4.50$ 	& $0.09$ 	& $3.70$ 	& $0.06$ \\
        ~	& ~	& $(4.48)$ 	& $(0.11)$ 		& $(4.45)$ 	& $(0.11)$ 	& $(2.99)$ 	& $(0.08)$ 	& $(3.05)$ 	& $(0.07)$	& $(2.93)$ 	& $(0.08)$ 	& $(3.09)$ 	& $(0.08)$\\
        ~ & \multirow{2}{*}{$150$px}	& $6.80$ 	& $0.10$	& $6.46$ 	& $0.09$ 	& $6.71$ 	& $0.11$	& $3.92$ 	& $0.05$ 	& $6.64$ 	& $0.11$ 	& $4.65$ 	& $0.07$ \\
        ~	& ~	& $(4.48)$ 	& $(0.10)$ 	& $(4.33)$ 	& $(0.09)$ 	& $(3.79)$ 	& $(0.08)$ 	& $(3.07)$ 	& $(0.06)$ 	& $(3.90)$ 	& $(0.08)$ 	& $(3.54)$ 	& $(0.07)$\\
        \cmidrule[0.25pt]{1-14}
        \multirow{6}{*}{\rotatebox[origin=c]{90}{Frame}} & \multirow{2}{*}{$20$px}	& $9.86$ 	& $0.18$ & $10.1$ 	& $0.20$ & $6.07$ 	& $0.16$ & $4.64$ 	& $0.12$  & $4.77$ 	& $0.13$ 	& $4.32$ 	& $0.11$ \\
        ~	& ~	& $(8.37)$ 	& $(0.20)$ & $(7.94)$ 	& $(0.21)$ & $(3.74)$ 	& $(0.14)$ 	& $(3.61)$ 	& $(0.15)$ & $(2.88)$ 	& $(0.12)$ 	& $(2.90)$ 	& $(0.11)$\\
        ~ & \multirow{2}{*}{$34$px}	& $8.92$ 	& $0.13$	& $8.63$ 	& $0.13$ & $6.71$ 	& $0.15$ & $4.50$ 	& $0.08$  & $5.68$ 	& $0.12$ 	& $4.85$ 	& $0.10$ \\
        ~	& ~	& $(6.60)$ 	& $(0.13)$ & $(6.52)$ 	& $(0.14)$	& $(4.04)$ 	& $(0.12)$	& $(2.96)$ 	& $(0.08)$ & $(3.25)$ 	& $(0.09)$ 	& $(3.44)$ 	& $(0.10)$\\
        ~ & \multirow{2}{*}{$58$px} &  $8.44$ 	& $0.12$ 	& $7.23$ 	& $0.09$ 	& $7.79$ 	& $0.14$ & $5.44$ 	& $0.08$  		& $7.02$ 	& $0.12$ 	& $5.78$ 	& $0.09$ \\
        ~	& ~	& $(5.72)$ 	& $(0.13)$ & $(4.94)$ 	& $(0.11)$	& $(4.17)$ 	& $(0.09)$ 	& $(3.89)$ 	& $(0.09)$ 	& $(3.90)$ 	& $(0.09)$ 	& $(3.79)$ 	& $(0.08)$\\
        \cmidrule[0.25pt]{1-14}
        \multirow{6}{*}{\rotatebox[origin=c]{90}{Random}} & \multirow{2}{*}{$17\%$}	& $8.11$ 	& $0.15$ & $7.41$ 	& $0.13$  & $5.20$ 	& $0.13$ & $4.43$ 	& $0.10$ & $4.59$ 	& $0.10$ 	& $3.81$ 	& $0.08$ \\
        ~	& ~	& $(5.63)$ 	& $(0.16)$ 	& $(4.63)$ 	& $(0.14)$ & $(3.14)$ 	& $(0.11)$ 	& $(3.30)$ 	& $(0.12)$ & $(2.65)$ 	& $(0.09)$ 	& $(2.92)$ 	& $(0.09)$\\
        ~ & \multirow{2}{*}{$28\%$} & $6.82$ 	& $0.10$ & $7.51$ & $0.11$ & $5.97$ 	& $0.12$ & $4.29$ 	& $0.07$ & $5.50$ 	& $0.11$ 	& $4.35$ 	& $0.07$ \\
        ~	& ~	& $(4.99)$ 	& $(0.12)$ 	& $(4.54)$ 	& $(0.11)$	& $(3.55)$ 	& $(0.10)$ & $(2.91)$ 	& $(0.07)$ & $(3.14)$ 	& $(0.09)$ 	& $(3.02)$ 	& $(0.08)$\\
        ~ & \multirow{2}{*}{$45\%$}	& $7.42$ 	& $0.10$ & $6.76$ 	& $0.09$ & $7.21$ 	& $0.12$ & $4.61$ 	& $0.06$ & $7.39$ 	& $0.12$ 	& $5.04$ 	& $0.07$ \\
        ~	& ~	& $(4.80)$ 	& $(0.11)$ & $(4.20)$ 	& $(0.09)$ & $(3.98)$ 	& $(0.09)$ 	& $(3.41)$ 	& $(0.07)$ 	& $(4.03)$ 	& $(0.08)$ 	& $(3.53)$ 	& $(0.07)$\\
\cmidrule[1pt]{1-14}
\end{tabular}
\vspace{-2em}
\label{tbl:exp_results_ldist}
\end{table*}

\textbf{Perturbation regions} In this study, we evaluate the use of three different perturbation regions, each with three different settings. These regions are (1) randomly selected pixels, (2) center square pixels, and (3) outer frame pixels. For (1), we randomly select $\{45\%, 28\%, 17\% \}$ of all pixels, where these percentages approximately correspond to a center square with a side length of $\{90, 120, 150\}$ pixels and an outer frame with a width of $\{20, 34, 58\}$ pixels, respectively. Thus, the number of selected pixels for all regions in each of the three different settings is virtually the same. Visual examples of the localization masks are provided in \cref{fig:pert_localiztion_masks} in \cref{appA}.

\textbf{Calculating $L_p$ distances} We calculate $L_p$ distances ($p = 0, 2, \infty$) between genuine images and their adversarial counterparts, similar to  calculations in~\citet{JSMA} and \citet{DBLP:journalsCarliniW17}. A detailed description of these calculations for our settings is also provided in \cref{appA}.


\section{Experiments}
\label{Experiments}
We first analyze model-to-model transferability (also called black-box transferability) for adversarial examples with localized perturbation. For each model-to-model pair, we generate $2,000$ adversarial examples that transfer from the source model to the target model. Using the same initial images as these adversarial examples do, we now apply perturbation to nine different regions (i.e., three regions, with each region coming with three different settings). In \cref{fig:localized_transferability}, we present the percentage of adversarial examples that transfer from model to model when localized perturbation is applied, as opposed to performing the adversarial attack without any localization constraints. We see that a large portion of adversarial examples maintains model-to-model transferability when perturbation is applied to local regions.


For the adversarial examples that maintain model-to-model transferability, \cref{tbl:exp_results_ldist} provides exhaustive details on the mean and standard deviation of the $L_2$ and $L_\infty$ properties of the produced adversarial examples. $L_0$ norms are omitted from this table because adversarial examples with regional perturbation almost always have reduced mean $L_0$ norms (\cref{fig:percentage_of_adv_with_less_lp_norm} and \cref{tbl:exp_results_l0} in \cref{appB}). Adversarial perturbation applied to the center square of an image reduces the mean $L_2$ norm while it increases the mean $L_{\infty}$ norm. However, with additional experiments, we discover that $43\%$ of the individual adversarial examples with localized perturbation have lower $L_{\infty}$ distances than their non-locally perturbed counterparts, showing that localized perturbation nevertheless reduces the $L_{\infty}$ norm for a large number of cases. The detailed breakdown of this analysis can be found in \cref{appB}.

Another important observation we make is the difference in perturbation for different regions. As can be seen, not all regions are equally important when it comes to manipulating the prediction of a DNN with adversarial perturbation. We clearly observe adversarial perturbation applied to the center square being more influential than perturbation in other regions. Surprisingly, applying perturbation to randomly selected pixels requires less distortion than applying it to the frame of an image, further highlighting the differences between \textit{important} and \textit{unimportant} regions. Allowing perturbation in a more condensed area versus a more expanded area provides different results for the center square region and the other two regions. Increasing the number of selected pixels in the center square region also increases the $L_2$ norm of the perturbation, while doing so for frame and random pixels reduces the aforementioned norm.

\section{Conclusion and Future Directions}
\label{Conclustion and Future Directions}
We have proposed a simple and general method for localizing perturbations generated by existing adversarial attacks to specific image regions. Our method is experimentally confirmed to be effective, maintaining high black-box transferability at distortion levels that are significantly lower than the distortion levels required by existing attacks. The reduction in the amount of perturbation achieved by our method raises the concern that existing adversarial defenses may be undermined, since these are usually designed to be effective only against non-local attacks requiring larger perturbation budgets. 

Our main priority for future work is (1) to investigate to what extent our localization method can fool state-of-the-art adversarial defenses as well as (2) to more precisely identify regions of importance where this localized perturbation can be made more effective, linking the observations made in this study to the interpretability of DNNs.

\bibliography{2020_04}
\bibliographystyle{icml2020}

\clearpage
\appendix
\section{Experimental Details}\label{appA}

\begin{figure}[h!]
\centering
\begin{tikzpicture}
\scriptsize
\centering
\def\sety1{0}
\node[align=center] at (0, \sety1) {Center square\\localization};
\node[inner sep=0pt] (a) at (2, \sety1)
    {\includegraphics[width=1.8cm]{localization_masks/c1.png}};
\node[align=center] at (2, \sety1-1.15) {$16.1\%$};
\node[inner sep=0pt] (b) at (4, \sety1)
    {\includegraphics[width=1.8cm]{localization_masks/c2.png}};
\node[align=center] at (4, \sety1-1.15) {$28.6\%$};
\node[inner sep=0pt] (c) at (6, \sety1)
    {\includegraphics[width=1.8cm]{localization_masks/c3.png}};
\node[align=center] at (6, \sety1-1.15) {$44.8\%$};

\def\sety1{-2.5}
\node[align=center] at (0, \sety1) {Random\\localization};
\node[inner sep=0pt] (russell) at (2, \sety1)
    {\includegraphics[width=1.8cm]{localization_masks/r1.png}};
\node[align=center] at (2, \sety1-1.15) {$17.0\%$};
\node[inner sep=0pt] (whitehead) at (4, \sety1)
    {\includegraphics[width=1.8cm]{localization_masks/r2.png}};
\node[align=center] at (4, \sety1-1.15) {$28.0\%$};
\node[inner sep=0pt] (whitehead) at (6, \sety1)
    {\includegraphics[width=1.8cm]{localization_masks/r3.png}};
\node[align=center] at (6, \sety1-1.15) {$45.0\%$};

\def\sety1{-5}
\node[align=center] at (0, \sety1) {Frame\\localization};
\node[inner sep=0pt] (russell) at (2, \sety1)
    {\includegraphics[width=1.8cm]{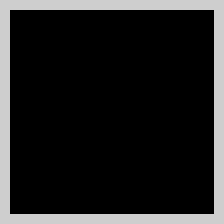}};
\node[align=center] at (2, \sety1-1.15) {$17.0\%$};
\node[inner sep=0pt] (whitehead) at (4, \sety1)
    {\includegraphics[width=1.8cm]{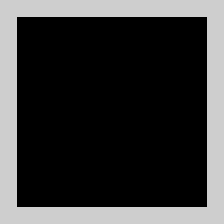}};
\node[align=center] at (4, \sety1-1.15) {$28.0\%$};
\node[inner sep=0pt] (whitehead) at (6, \sety1)
    {\includegraphics[width=1.8cm]{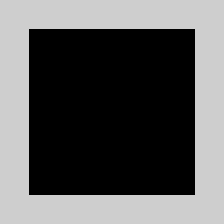}};
\node[align=center] at (6, \sety1-1.15) {$45.0\%$};
\end{tikzpicture}
\vspace{-1em}
\caption{The localization masks used in this study. The given percentages correspond to the number of selected pixels compared to the total number of available pixels.}
\label{fig:pert_localiztion_masks}
\end{figure}

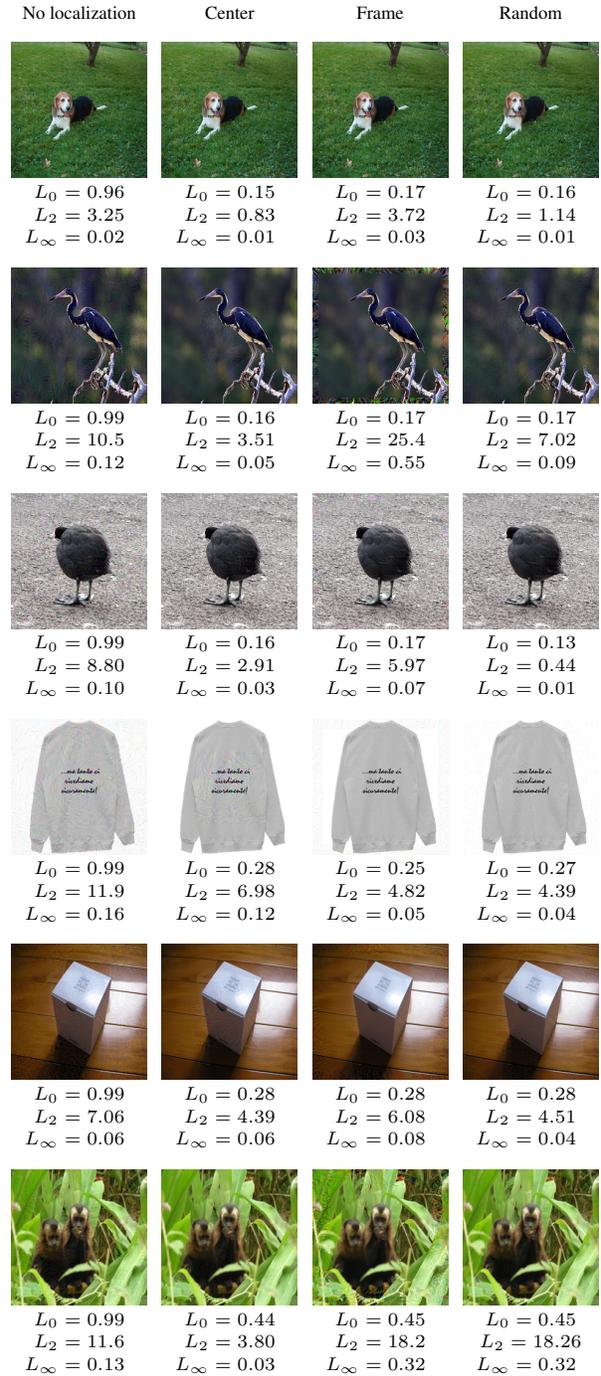
\begin{figure}[ht!]
\centering
\begin{tikzpicture}
\scriptsize
\centering
\def\sety1{1.3}
\node[align=center] at (0, \sety1) {No localization};
\node[align=center] at (2, \sety1) {Center};
\node[align=center] at (4, \sety1) {Frame};
\node[align=center] at (6, \sety1) {Random};

\def\sety1{0}
\node[inner sep=0pt] (a) at (0, \sety1)
    {\includegraphics[width=1.8cm]{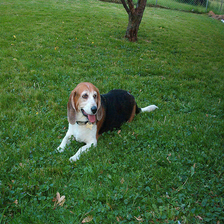}};
\node[align=center] at (0, \sety1-1.1) {$L_0=0.96$};
\node[align=center] at (0, \sety1-1.4) {$L_2=3.25$};
\node[align=center] at (0-0.06, \sety1-1.7) {$L_{\infty}=0.02$};
\node[inner sep=0pt] (a) at (2, \sety1)
    {\includegraphics[width=1.8cm]{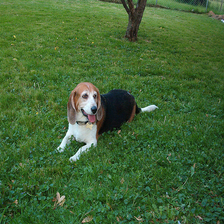}};
\node[align=center] at (2, \sety1-1.1) {$L_0=0.15$};
\node[align=center] at (2, \sety1-1.4) {$L_2=0.83$};
\node[align=center] at (2-0.06, \sety1-1.7) {$L_{\infty}=0.01$};
\node[inner sep=0pt] (b) at (4, \sety1)
    {\includegraphics[width=1.8cm]{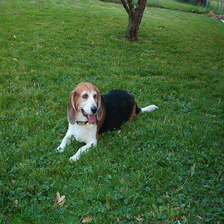}};
\node[align=center] at (4, \sety1-1.1) {$L_0=0.17$};
\node[align=center] at (4, \sety1-1.4) {$L_2=3.72$};
\node[align=center] at (4-0.06, \sety1-1.7) {$L_{\infty}=0.03$};
\node[inner sep=0pt] (c) at (6, \sety1)
    {\includegraphics[width=1.8cm]{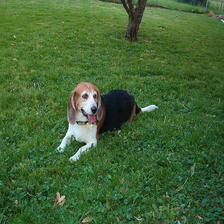}};
\node[align=center] at (6, \sety1-1.1) {$L_0=0.16$};
\node[align=center] at (6, \sety1-1.4) {$L_2=1.14$};
\node[align=center] at (6-0.06, \sety1-1.7) {$L_{\infty}=0.01$};

\def\sety1{-3}
\node[inner sep=0pt] (a) at (0, \sety1)
    {\includegraphics[width=1.8cm]{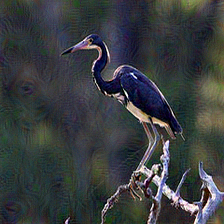}};
\node[align=center] at (0, \sety1-1.1) {$L_0=0.99$};
\node[align=center] at (0, \sety1-1.4) {$L_2=10.5$};
\node[align=center] at (0-0.06, \sety1-1.7) {$L_{\infty}=0.12$};
\node[inner sep=0pt] (a) at (2, \sety1)
    {\includegraphics[width=1.8cm]{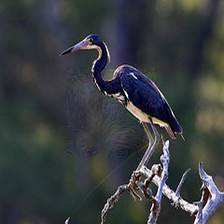}};
\node[align=center] at (2, \sety1-1.1) {$L_0=0.16$};
\node[align=center] at (2, \sety1-1.4) {$L_2=3.51$};
\node[align=center] at (2-0.06, \sety1-1.7) {$L_{\infty}=0.05$};
\node[inner sep=0pt] (b) at (4, \sety1)
    {\includegraphics[width=1.8cm]{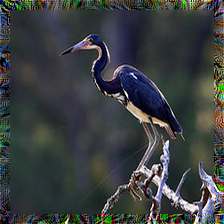}};
\node[align=center] at (4, \sety1-1.1) {$L_0=0.17$};
\node[align=center] at (4, \sety1-1.4) {$L_2=25.4$};
\node[align=center] at (4-0.06, \sety1-1.7) {$L_{\infty}=0.55$};
\node[inner sep=0pt] (c) at (6, \sety1)
    {\includegraphics[width=1.8cm]{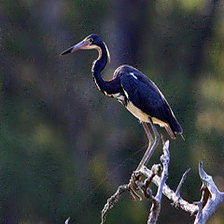}};
\node[align=center] at (6, \sety1-1.1) {$L_0=0.17$};
\node[align=center] at (6, \sety1-1.4) {$L_2=7.02$};
\node[align=center] at (6-0.06, \sety1-1.7) {$L_{\infty}=0.09$};

\def\sety1{-6}
\node[inner sep=0pt] (a) at (0, \sety1)
    {\includegraphics[width=1.8cm]{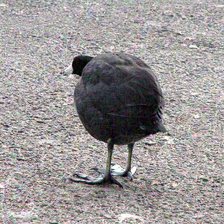}};
\node[align=center] at (0, \sety1-1.1) {$L_0=0.99$};
\node[align=center] at (0, \sety1-1.4) {$L_2=8.80$};
\node[align=center] at (0-0.06, \sety1-1.7) {$L_{\infty}=0.10$};
\node[inner sep=0pt] (a) at (2, \sety1)
    {\includegraphics[width=1.8cm]{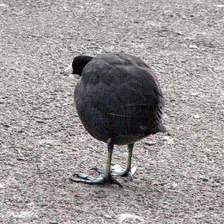}};
\node[align=center] at (2, \sety1-1.1) {$L_0=0.16$};
\node[align=center] at (2, \sety1-1.4) {$L_2=2.91$};
\node[align=center] at (2-0.06, \sety1-1.7) {$L_{\infty}=0.03$};
\node[inner sep=0pt] (b) at (4, \sety1)
    {\includegraphics[width=1.8cm]{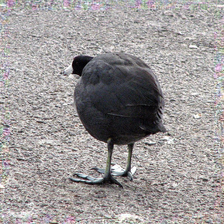}};
\node[align=center] at (4, \sety1-1.1) {$L_0=0.17$};
\node[align=center] at (4, \sety1-1.4) {$L_2=5.97$};
\node[align=center] at (4-0.06, \sety1-1.7) {$L_{\infty}=0.07$};
\node[inner sep=0pt] (c) at (6, \sety1)
    {\includegraphics[width=1.8cm]{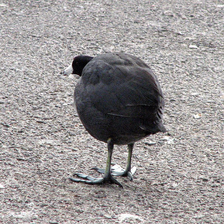}};
\node[align=center] at (6, \sety1-1.1) {$L_0=0.13$};
\node[align=center] at (6, \sety1-1.4) {$L_2=0.44$};
\node[align=center] at (6-0.06, \sety1-1.7) {$L_{\infty}=0.01$};

\def\sety1{-9}
\node[inner sep=0pt] (a) at (0, \sety1)
    {\includegraphics[width=1.8cm]{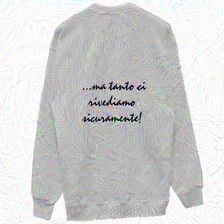}};
\node[align=center] at (0, \sety1-1.1) {$L_0=0.99$};
\node[align=center] at (0, \sety1-1.4) {$L_2=11.9$};
\node[align=center] at (0-0.06, \sety1-1.7) {$L_{\infty}=0.16$};
\node[inner sep=0pt] (a) at (2, \sety1)
    {\includegraphics[width=1.8cm]{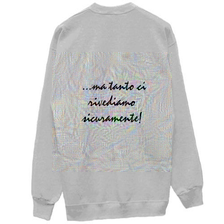}};
\node[align=center] at (2, \sety1-1.1) {$L_0=0.28$};
\node[align=center] at (2, \sety1-1.4) {$L_2=6.98$};
\node[align=center] at (2-0.06, \sety1-1.7) {$L_{\infty}=0.12$};
\node[inner sep=0pt] (b) at (4, \sety1)
    {\includegraphics[width=1.8cm]{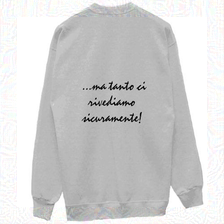}};
\node[align=center] at (4, \sety1-1.1) {$L_0=0.25$};
\node[align=center] at (4, \sety1-1.4) {$L_2=4.82$};
\node[align=center] at (4-0.06, \sety1-1.7) {$L_{\infty}=0.05$};
\node[inner sep=0pt] (c) at (6, \sety1)
    {\includegraphics[width=1.8cm]{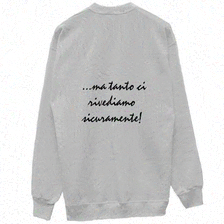}};
\node[align=center] at (6, \sety1-1.1) {$L_0=0.27$};
\node[align=center] at (6, \sety1-1.4) {$L_2=4.39$};
\node[align=center] at (6-0.06, \sety1-1.7) {$L_{\infty}=0.04$};

\def\sety1{-12}
\node[inner sep=0pt] (a) at (0, \sety1)
    {\includegraphics[width=1.8cm]{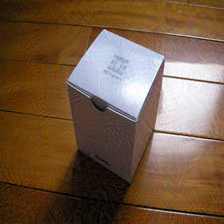}};
\node[align=center] at (0, \sety1-1.1) {$L_0=0.99$};
\node[align=center] at (0, \sety1-1.4) {$L_2=7.06$};
\node[align=center] at (0-0.06, \sety1-1.7) {$L_{\infty}=0.06$};
\node[inner sep=0pt] (a) at (2, \sety1)
    {\includegraphics[width=1.8cm]{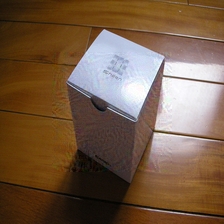}};
\node[align=center] at (2, \sety1-1.1) {$L_0=0.28$};
\node[align=center] at (2, \sety1-1.4) {$L_2=4.39$};
\node[align=center] at (2-0.06, \sety1-1.7) {$L_{\infty}=0.06$};
\node[inner sep=0pt] (b) at (4, \sety1)
    {\includegraphics[width=1.8cm]{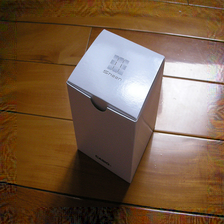}};
\node[align=center] at (4, \sety1-1.1) {$L_0=0.28$};
\node[align=center] at (4, \sety1-1.4) {$L_2=6.08$};
\node[align=center] at (4-0.06, \sety1-1.7) {$L_{\infty}=0.08$};
\node[inner sep=0pt] (c) at (6, \sety1)
    {\includegraphics[width=1.8cm]{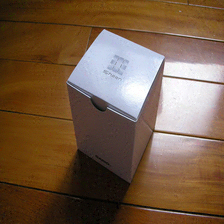}};
\node[align=center] at (6, \sety1-1.1) {$L_0=0.28$};
\node[align=center] at (6, \sety1-1.4) {$L_2=4.51$};
\node[align=center] at (6-0.06, \sety1-1.7) {$L_{\infty}=0.04$};

\def\sety1{-15}
\node[inner sep=0pt] (a) at (0, \sety1)
    {\includegraphics[width=1.8cm]{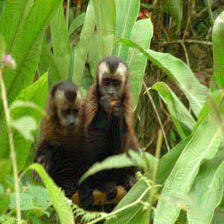}};
\node[align=center] at (0, \sety1-1.1) {$L_0=0.99$};
\node[align=center] at (0, \sety1-1.4) {$L_2=11.6$};
\node[align=center] at (0-0.06, \sety1-1.7) {$L_{\infty}=0.13$};
\node[inner sep=0pt] (a) at (2, \sety1)
    {\includegraphics[width=1.8cm]{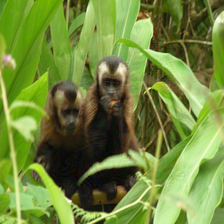}};
\node[align=center] at (2, \sety1-1.1) {$L_0=0.44$};
\node[align=center] at (2, \sety1-1.4) {$L_2=3.80$};
\node[align=center] at (2-0.06, \sety1-1.7) {$L_{\infty}=0.03$};
\node[inner sep=0pt] (b) at (4, \sety1)
    {\includegraphics[width=1.8cm]{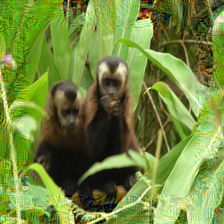}};
\node[align=center] at (4, \sety1-1.1) {$L_0=0.45$};
\node[align=center] at (4, \sety1-1.4) {$L_2=18.2$};
\node[align=center] at (4-0.06, \sety1-1.7) {$L_{\infty}=0.32$};
\node[inner sep=0pt] (c) at (6, \sety1)
    {\includegraphics[width=1.8cm]{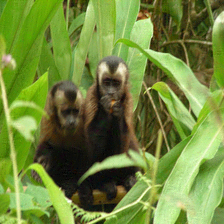}};
\node[align=center] at (6, \sety1-1.1) {$L_0=0.45$};
\node[align=center] at (6, \sety1-1.4) {$L_2=18.26$};
\node[align=center] at (6-0.06, \sety1-1.7) {$L_{\infty}=0.32$};

\end{tikzpicture}
\vspace{-1em}
\caption{$L_0$, $L_2$, and $L_{\infty}$ distances between the initial images and their adversarial counterparts for the adversarial examples that originate from the same initial image but that were perturbed using different localization methods. All of the adversarial examples successfully transfer to models they are not originated from.}
\label{fig:additional_examples}
\vspace{-1em}
\end{figure}

\begin{figure*}[t!]
\setlength\unitlength{1cm}
\centering
\begin{minipage}{.3\textwidth}
\centering
\begin{picture}(4,4)
\multiput(0.1,0.1)(0,1){4}{\line(1,0){3}}
\multiput(0.1,0.1)(1,0){4}{\line(0,1){3}}
\begin{scriptsize}
\put(0.3,0.55){$99\%$}
\put(1.3,0.55){$99\%$}
\put(2.45,0.55){\textemdash}

\put(0.3,1.55){$99\%$}
\put(1.45,1.55){\textemdash}
\put(2.3,1.55){$99\%$}

\put(0.45,2.55){\textemdash}
\put(1.3,2.55){$99\%$}
\put(2.3,2.55){$99\%$}

\put(-0.3,2.3){\rotatebox{90}{AlexNet}}
\put(-0.3,1.25){\rotatebox{90}{VGG-16}}
\put(-0.3,0.05){\rotatebox{90}{ResNet-50}}

\put(0.12,3.3){AlexNet}
\put(1.12,3.3){VGG-16}
\put(2.12,3.3){ResNet-50}
\end{scriptsize}
\put (-0.75,0.75){\rotatebox{90}{Source Model}}
\put(0.75,3.75){Target Model}
\put(-0.5,-0.5){(a) $17\%$ of pixels selected}
\end{picture}
\end{minipage}
\begin{minipage}{.3\textwidth}
\centering
\begin{picture}(4,4)
\multiput(0.1,0.1)(0,1){4}{\line(1,0){3}}
\multiput(0.1,0.1)(1,0){4}{\line(0,1){3}}
\begin{scriptsize}
\put(0.3,0.55){$99\%$}
\put(1.3,0.55){$99\%$}
\put(2.45,0.55){\textemdash}

\put(0.3,1.55){$99\%$}
\put(1.45,1.55){\textemdash}
\put(2.3,1.55){$99\%$}

\put(0.45,2.55){\textemdash}
\put(1.3,2.55){$99\%$}
\put(2.3,2.55){$99\%$}

\put(-0.3,2.3){\rotatebox{90}{AlexNet}}
\put(-0.3,1.25){\rotatebox{90}{VGG-16}}
\put(-0.3,0.05){\rotatebox{90}{ResNet-50}}

\put(0.12,3.3){AlexNet}
\put(1.12,3.3){VGG-16}
\put(2.12,3.3){ResNet-50}
\end{scriptsize}
\put (-0.75,0.75){\rotatebox{90}{Source Model}}
\put(0.75,3.75){Target Model}
\put(-0.5,-0.5){(b) $28\%$ of pixels selected}
\end{picture}
\end{minipage}
\begin{minipage}{.3\textwidth}
\centering
\begin{picture}(4,4)
\multiput(0.1,0.1)(0,1){4}{\line(1,0){3}}
\multiput(0.1,0.1)(1,0){4}{\line(0,1){3}}
\begin{scriptsize}
\put(0.3,0.55){$99\%$}
\put(1.3,0.55){$99\%$}
\put(2.45,0.55){\textemdash}

\put(0.3,1.55){$99\%$}
\put(1.45,1.55){\textemdash}
\put(2.3,1.55){$99\%$}

\put(0.45,2.55){\textemdash}
\put(1.3,2.55){$99\%$}
\put(2.3,2.55){$99\%$}

\put(-0.3,2.3){\rotatebox{90}{AlexNet}}
\put(-0.3,1.25){\rotatebox{90}{VGG-16}}
\put(-0.3,0.05){\rotatebox{90}{ResNet-50}}

\put(0.12,3.3){AlexNet}
\put(1.12,3.3){VGG-16}
\put(2.12,3.3){ResNet-50}
\end{scriptsize}
\put (-0.75,0.75){\rotatebox{90}{Source Model}}
\put(0.75,3.75){Target Model}
\put(-0.5,-0.5){(c) $45\%$ of pixels selected}
\end{picture}
\end{minipage}
\vspace{4em}

\begin{minipage}{.3\textwidth}
\centering
\begin{picture}(4,4)
\multiput(0.1,0.1)(0,1){4}{\line(1,0){3}}
\multiput(0.1,0.1)(1,0){4}{\line(0,1){3}}
\begin{scriptsize}
\put(0.3,0.55){$82\%$}
\put(1.3,0.55){$85\%$}
\put(2.45,0.55){\textemdash}

\put(0.3,1.55){$77\%$}
\put(1.45,1.55){\textemdash}
\put(2.3,1.55){$84\%$}

\put(0.45,2.55){\textemdash}
\put(1.3,2.55){$77\%$}
\put(2.3,2.55){$81\%$}

\put(-0.3,2.3){\rotatebox{90}{AlexNet}}
\put(-0.3,1.25){\rotatebox{90}{VGG-16}}
\put(-0.3,0.05){\rotatebox{90}{ResNet-50}}

\put(0.12,3.3){AlexNet}
\put(1.12,3.3){VGG-16}
\put(2.12,3.3){ResNet-50}
\end{scriptsize}
\put (-0.75,0.75){\rotatebox{90}{Source Model}}
\put(0.75,3.75){Target Model}
\put(-0.5,-0.5){(a) $17\%$ of pixels selected}
\end{picture}
\end{minipage}
\begin{minipage}{.3\textwidth}
\centering
\begin{picture}(4,4)
\multiput(0.1,0.1)(0,1){4}{\line(1,0){3}}
\multiput(0.1,0.1)(1,0){4}{\line(0,1){3}}
\begin{scriptsize}
\put(0.3,0.55){$80\%$}
\put(1.3,0.55){$79\%$}
\put(2.45,0.55){\textemdash}

\put(0.3,1.55){$78\%$}
\put(1.45,1.55){\textemdash}
\put(2.3,1.55){$79\%$}

\put(0.45,2.55){\textemdash}
\put(1.3,2.55){$80\%$}
\put(2.3,2.55){$76\%$}

\put(-0.3,2.3){\rotatebox{90}{AlexNet}}
\put(-0.3,1.25){\rotatebox{90}{VGG-16}}
\put(-0.3,0.05){\rotatebox{90}{ResNet-50}}

\put(0.12,3.3){AlexNet}
\put(1.12,3.3){VGG-16}
\put(2.12,3.3){ResNet-50}
\end{scriptsize}
\put (-0.75,0.75){\rotatebox{90}{Source Model}}
\put(0.75,3.75){Target Model}
\put(-0.5,-0.5){(b) $28\%$ of pixels selected}
\end{picture}
\end{minipage}
\begin{minipage}{.3\textwidth}
\centering
\begin{picture}(4,4)
\multiput(0.1,0.1)(0,1){4}{\line(1,0){3}}
\multiput(0.1,0.1)(1,0){4}{\line(0,1){3}}
\begin{scriptsize}
\put(0.3,0.55){$77\%$}
\put(1.3,0.55){$76\%$}
\put(2.45,0.55){\textemdash}

\put(0.3,1.55){$74\%$}
\put(1.45,1.55){\textemdash}
\put(2.3,1.55){$68\%$}

\put(0.45,2.55){\textemdash}
\put(1.3,2.55){$80\%$}
\put(2.3,2.55){$74\%$}

\put(-0.3,2.3){\rotatebox{90}{AlexNet}}
\put(-0.3,1.25){\rotatebox{90}{VGG-16}}
\put(-0.3,0.05){\rotatebox{90}{ResNet-50}}

\put(0.12,3.3){AlexNet}
\put(1.12,3.3){VGG-16}
\put(2.12,3.3){ResNet-50}
\end{scriptsize}
\put (-0.75,0.75){\rotatebox{90}{Source Model}}
\put(0.75,3.75){Target Model}
\put(-0.5,-0.5){(c) $45\%$ of pixels selected}
\end{picture}
\end{minipage}
\vspace{4em}

\begin{minipage}{.3\textwidth}
\centering
\begin{picture}(4,4)
\multiput(0.1,0.1)(0,1){4}{\line(1,0){3}}
\multiput(0.1,0.1)(1,0){4}{\line(0,1){3}}
\begin{scriptsize}
\put(0.3,0.55){$32\%$}
\put(1.3,0.55){$40\%$}
\put(2.45,0.55){\textemdash}

\put(0.3,1.55){$28\%$}
\put(1.45,1.55){\textemdash}
\put(2.3,1.55){$41\%$}

\put(0.45,2.55){\textemdash}
\put(1.3,2.55){$52\%$}
\put(2.3,2.55){$43\%$}

\put(-0.3,2.3){\rotatebox{90}{AlexNet}}
\put(-0.3,1.25){\rotatebox{90}{VGG-16}}
\put(-0.3,0.05){\rotatebox{90}{ResNet-50}}

\put(0.12,3.3){AlexNet}
\put(1.12,3.3){VGG-16}
\put(2.12,3.3){ResNet-50}
\end{scriptsize}
\put (-0.75,0.75){\rotatebox{90}{Source Model}}
\put(0.75,3.75){Target Model}
\put(-0.5,-0.5){(a) $17\%$ of pixels selected}
\end{picture}
\end{minipage}
\begin{minipage}{.3\textwidth}
\centering
\begin{picture}(4,4)
\multiput(0.1,0.1)(0,1){4}{\line(1,0){3}}
\multiput(0.1,0.1)(1,0){4}{\line(0,1){3}}
\begin{scriptsize}
\put(0.3,0.55){$28\%$}
\put(1.3,0.55){$49\%$}
\put(2.45,0.55){\textemdash}

\put(0.3,1.55){$34\%$}
\put(1.45,1.55){\textemdash}
\put(2.3,1.55){$49\%$}

\put(0.45,2.55){\textemdash}
\put(1.3,2.55){$49\%$}
\put(2.3,2.55){$47\%$}

\put(-0.3,2.3){\rotatebox{90}{AlexNet}}
\put(-0.3,1.25){\rotatebox{90}{VGG-16}}
\put(-0.3,0.05){\rotatebox{90}{ResNet-50}}

\put(0.12,3.3){AlexNet}
\put(1.12,3.3){VGG-16}
\put(2.12,3.3){ResNet-50}
\end{scriptsize}
\put (-0.75,0.75){\rotatebox{90}{Source Model}}
\put(0.75,3.75){Target Model}
\put(-0.5,-0.5){(b) $28\%$ of pixels selected}
\end{picture}
\end{minipage}
\begin{minipage}{.3\textwidth}
\centering
\begin{picture}(4,4)
\multiput(0.1,0.1)(0,1){4}{\line(1,0){3}}
\multiput(0.1,0.1)(1,0){4}{\line(0,1){3}}
\begin{scriptsize}
\put(0.3,0.55){$34\%$}
\put(1.3,0.55){$45\%$}
\put(2.45,0.55){\textemdash}

\put(0.3,1.55){$34\%$}
\put(1.45,1.55){\textemdash}
\put(2.3,1.55){$50\%$}

\put(0.45,2.55){\textemdash}
\put(1.3,2.55){$62\%$}
\put(2.3,2.55){$55\%$}

\put(-0.3,2.3){\rotatebox{90}{AlexNet}}
\put(-0.3,1.25){\rotatebox{90}{VGG-16}}
\put(-0.3,0.05){\rotatebox{90}{ResNet-50}}

\put(0.12,3.3){AlexNet}
\put(1.12,3.3){VGG-16}
\put(2.12,3.3){ResNet-50}
\end{scriptsize}
\put (-0.75,0.75){\rotatebox{90}{Source Model}}
\put(0.75,3.75){Target Model}
\put(-0.5,-0.5){(c) $45\%$ of pixels selected}
\end{picture}
\end{minipage}
\vspace{2em}
\caption{The percentage of adversarial examples with regional perturbation that have less perturbation in terms of $L_0$ norm (top),  $L_2$ norm (middle), and $L_{\infty}$ norm (bottom) compared to their counterparts with ``global'' perturbation. Percentages are calculated based on the adversarial examples with localized perturbation that transfer from source model to target model.}
\label{fig:percentage_of_adv_with_less_lp_norm}
\end{figure*}
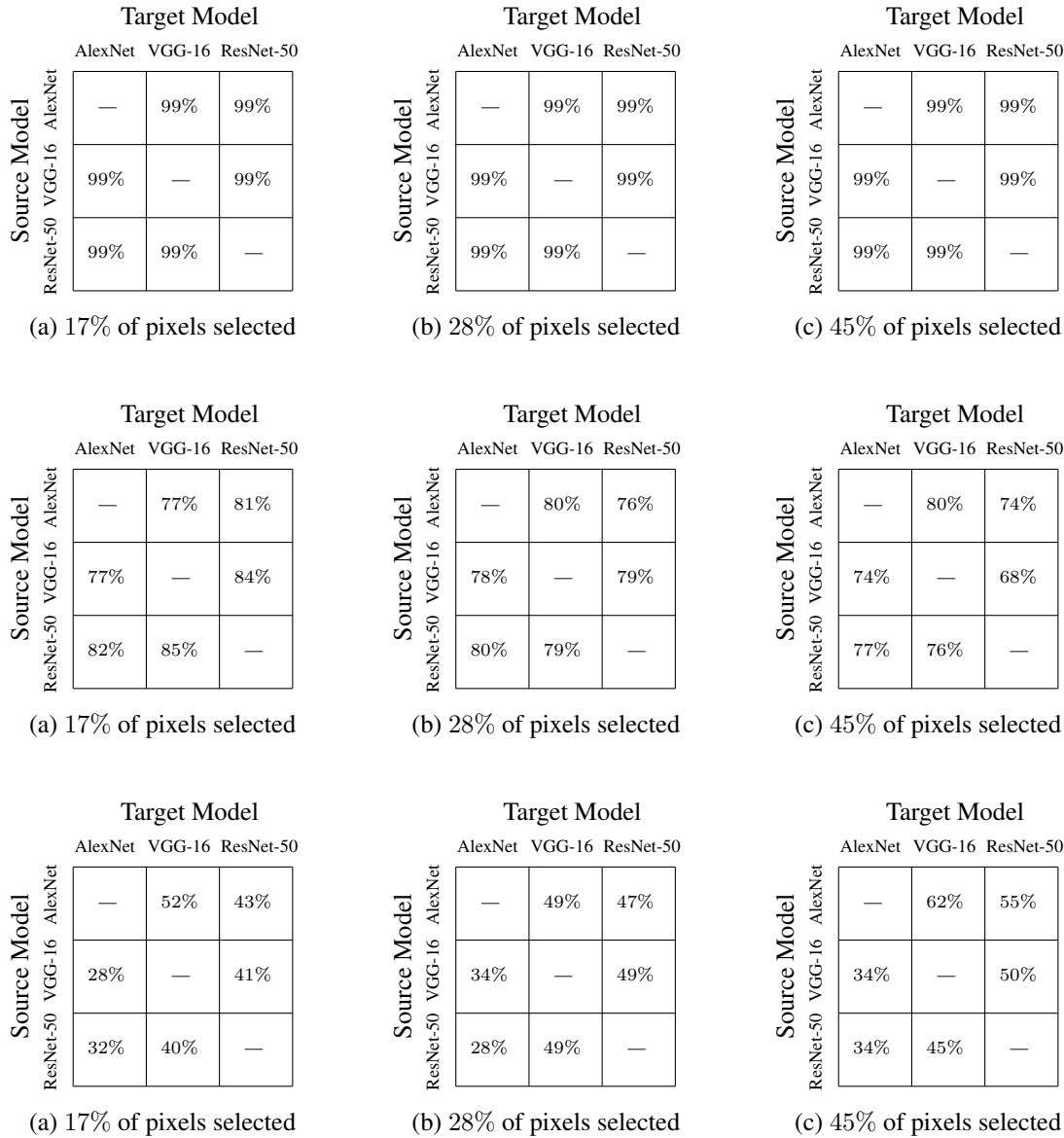   

\textbf{Perturbation regions} In Figure~\ref{fig:pert_localiztion_masks}, we provide visualizations of the selected localization masks, where the given percentages correspond to the proportion of selected pixels out of all available pixels. 

Note that for the perturbation localized on the image frame, unlike~\citet{zajac2019adversarial_frame}, we do not expand the size of the image. We simply exercise the perturbation on the selected outermost pixels.

\textbf{Calculating $L_p$ distances} Between initial images of size $224 \times 224$ and their adversarial counterparts, we calculate $L_0$, $L_2$, and $L_{\infty}$ distances as follows:
\begin{align}
    L_{0}(\bm{X}, \bm{\tilde{X}}) & = \frac{\sum_{i}^{224} \sum_{j}^{224} \mathds{1}_{\{\bm{X}_{i,j} - \bm{\tilde{X}}_{i,j} \neq 0\}}}{224 \times 224} \,, \\
    L_{2}(\bm{X}, \bm{\tilde{X}}) & = ||\bm{X} - \bm{\tilde{X}}||_2 \,,\\
    L_{\infty}(\bm{X}, \bm{\tilde{X}}) & = \max(|\bm{X} - \bm{\tilde{X}}|) \,,
\end{align}
where $\bm{X}$ and $\bm{\tilde X}$ represent an initial image and its adversarial counterpart, respectively. In this framework, an $L_{\infty}$ norm of $1$ means that the added perturbation changed a pixel from black to white (i.e., $0$ to $255$), or vice versa. An $L_0$ norm of $1$ means all pixels are modified by the adversarial perturbation.

\newpage\phantom{-}
\newpage
\section{Additional Experimental Results}\label{appB}
In Figure~\ref{fig:additional_examples}, we provide a number of qualitative examples, showing the $L_0$, $L_2$, and $L_{\infty}$ norms of adversarial perturbation generated using various localization settings. All of the examples presented in Figure~\ref{fig:additional_examples} are generated using AlexNet and transfer to ResNet-50.

For the experiments discussed in the main paper, Figure~\ref{fig:percentage_of_adv_with_less_lp_norm} provides the percentage of adversarial examples that have lower $L_p$ norm than their counterparts generated with  ``global'' perturbation. Our experiments show that regional perturbation almost always leads to lower $L_0$ norms compared to non-regional perturbation, whereas in the case of $L_2$ and $L_{\infty}$ norms, this depends on the initial image-target class combination.

For the sake of completeness, In \cref{tbl:exp_results_l0} we provide the exhaustive details of $L_0$ norms of  adversarial perturbations for the experiment described in the main paper. Since the perturbation region is what is controlled in this experiment, the resulting perturbations have much less $L_0$ deviation compared to $L_2$ or $L_{\infty}$.

\begin{table*}[t!]
\centering
\caption{Mean (standard deviation) $L_0$ distances calculated between genuine images and their adversarial counterparts, with the adversarial counterparts created by localization of perturbation (see the first column). Adversarial examples are created from the source models listed in the first row and transfer to the target models listed in the second row.}
\tiny
\begin{tabular}{llcccccc}
	\cmidrule[1pt]{1-8}
	 \multirow{3}{*}{\rotatebox[origin=c]{90}{Localization}} & Source: & \multicolumn{2}{c}{AlexNet} & \multicolumn{2}{c}{VGG-16} & \multicolumn{2}{c}{ResNet-50}\\ 
    \cmidrule[0.25pt]{3-8}
    ~ & Target: & VGG-16 & ResNet-50 & AlexNet & ResNet-50 & AlexNet & VGG-16 \\ 
    \cmidrule[0.25pt]{3-8}
    ~ & Norm: & \multicolumn{2}{c}{$L_0$} & \multicolumn{2}{c}{$L_0$}  & \multicolumn{2}{c}{$L_0$}\\ 
        \cmidrule[0.5pt]{1-8} 
        \multicolumn{2}{c}{\multirow{2}{*}{No Localization}} & $0.93$ 	& $0.94$	& $0.90$ 	& $0.84$	& $0.89$ 	& $0.83$ 	 \\
        ~	& ~	&  $(0.08)$ 	& $(0.07)$ 	& $(0.07)$ 	& $(0.11)$ & $(0.10)$ & $(0.10)$ \\
        \cmidrule[0.25pt]{1-8}
        \multirow{6}{*}{\rotatebox[origin=c]{90}{Center}} & \multirow{2}{*}{$90$px}	& $0.15$ 	& $0.15$ 	& $0.15$ 	& $0.15$ 	& $0.15$ 	& $0.14$ \\
        ~	& ~	& $(0.01)$ 	& $(0.01)$ 	& $(0.01)$ 	& $(0.01)$ & $(0.01)$ & $(0.01)$ \\
        ~ & \multirow{2}{*}{$120$px}	& $0.27$ 	& $0.28$ 	& $0.27$  & $0.26$ & $0.27$ & $0.26$   \\
        ~	& ~	& $(0.01)$ 	& $(0.01)$ 		& $(0.01)$ & $(0.03)$ & $(0.01)$ 	& $(0.03)$ 	\\
        ~ & \multirow{2}{*}{$150$px}	& $0.43$ 	& $0.43$	& $0.43$ & $0.40$ & $0.43$ & $0.41$  \\
        ~	& ~	& $(0.02)$ & $(0.02)$ & $(0.02)$ & $(0.04)$ & $(0.02)$ & $(0.04)$ \\
        \cmidrule[0.25pt]{1-8}
        \multirow{6}{*}{\rotatebox[origin=c]{90}{Frame}} & \multirow{2}{*}{$20$px}	& $0.16$ & $0.16$  & $0.16$  & $0.16$  & $0.16$  & $0.16$ \\
        ~	& ~	& $(0.01)$ & $(0.01)$ & $(0.01)$ & $(0.01)$ & $(0.01)$ & $(0.01)$ 	\\
        ~ & \multirow{2}{*}{$34$px}	& $0.27$ 	& $0.27$	& $0.27$ 	& $0.26$ & $0.27$ & $0.26$ 	\\
        ~	& ~	& $(0.01)$ 	& $(0.01)$& $(0.01)$& $(0.02)$& $(0.01)$& $(0.02)$  \\
        ~ & \multirow{2}{*}{$58$px} &  $0.43$ &  $0.43$ &  $0.43$ &  $0.42$ &  $0.43$ &  $0.42$   \\
        ~	& ~	& $(0.02)$ & $(0.02)$& $(0.02)$& $(0.04)$& $(0.04)$& $(0.03)$	\\
        \cmidrule[0.25pt]{1-8}
        \multirow{6}{*}{\rotatebox[origin=c]{90}{Random}} & \multirow{2}{*}{$17\%$}	& $0.16$ 	& $0.16$ & $0.16$ 	& $0.16$ & $0.16$  & $0.15$  \\
        ~	& ~	& $(0.01)$ & $(0.01)$& $(0.01)$& $(0.01)$& $(0.01)$& $(0.01)$ \\
        ~ & \multirow{2}{*}{$28\%$} & $0.28$ & $0.28$ & $0.28$ & $0.27$ & $0.28$ & $0.27$  \\
        ~	& ~	& $(0.01)$ 	& $(0.01)$& $(0.01)$& $(0.02)$& $(0.01)$& $(0.02)$ \\
        ~ & \multirow{2}{*}{$45\%$}	& $0.44$ & $0.43$ & $0.43$ & $0.41$ & $0.43$ & $0.41$  \\
        ~	& ~	& $(4.80)$ 	& $(0.11)$ & $(4.20)$ 	& $(0.09)$ & $(3.98)$ 	& $(0.09)$ 	\\
\cmidrule[1pt]{1-8}
\end{tabular}
\vspace{-2em}
\label{tbl:exp_results_l0}
\end{table*}

\end{document}